\documentclass[10pt]{article} 
\usepackage[preprint]{tmlr}

\usepackage{amsmath,amsfonts,bm}

\def\eqref#1{equation~\ref{#1}}

\def\1{\bm{1}}

\DeclareMathAlphabet{\mathsfit}{\encodingdefault}{\sfdefault}{m}{sl}
\SetMathAlphabet{\mathsfit}{bold}{\encodingdefault}{\sfdefault}{bx}{n}

\usepackage{xcolor}          
\definecolor{accent}{HTML}{1F4E79}   
\definecolor{extern}{HTML}{7D4E2D}   

\usepackage{hyperref}
\hypersetup{
  colorlinks   = true,
  allcolors    = accent,
  urlcolor     = extern,
  linktoc      = all,
  breaklinks   = true,
  bookmarksnumbered = true,
}
\usepackage[nameinlink,capitalise]{cleveref}  
\usepackage{xurl}            

\usepackage{url}
\usepackage{graphicx}
\usepackage{float}
\usepackage{array}

\title{Prior Choice and the Cold Posterior Effect in Practical Bayesian Neural Networks}

\author{\name Joshua Cheng \email jc4120@princeton.edu \\
      \addr Department of Mathematics\\
      Princeton University
      \AND
      \name James Odgers \email james.odgers@helmholtz-munich.de \\
      \addr University of Technology Nuremberg \\ Munich Center for Machine Learning (MCML) \\  Helmholtz AI
      \AND
      \name Vincent Fortuin \email vincent.fortuin@utn.de\\
      \addr University of Technology Nuremberg \\ Munich Center for Machine Learning (MCML) \\  Helmholtz AI}

\def\month{MM}  
\def\year{YYYY} 
\def\openreview{\url{https://openreview.net/forum?id=XXXX}} 

\begin{document}

\maketitle

\begin{abstract}
Bayesian Deep Learning (BDL) has previously been observed to suffer from a Cold Posterior Effect (CPE), where artificially reducing the uncertainty of the posterior improves predictive performance. In this work we investigate whether the choice of approximate inference algorithm, which is typically needed to make BDL practical, affects the strength of the CPE. In particular, we empirically study how prior choice interacts with a variety of practical approximate inference methods to produce or reduce the cold posterior effect. Across the settings studied, we observe that inference method and posterior temperature usually have a larger and more consistent effect on performance than the choice among simple prior families. We also find that cold posterior behavior is strongly dependent on the approximate inference technique, with mean-field variational methods showing a pronounced CPE, the Laplace approximation exhibiting varying behavior, and function-space particle methods showing clearer temperature sensitivity than their parameter-space counterparts. These results suggest that for practitioners using scalable approximate Bayesian inference, inference method selection and validation-based temperature tuning should often be prioritized before extensive tuning of simple prior families.
\end{abstract}

\section{Introduction}

Bayesian neural networks provide a principled framework for uncertainty-aware deep learning by placing a probability distribution over model parameters and updating these distributions using Bayes' rule \citep{neal1992bayesian}. While this Bayesian formulation formalizes uncertainty quantification by allowing neural networks to represent posterior uncertainty rather than only a point estimate, in practice, the theoretical framework does not always match observations. In particular, although standard Bayesian theory corresponds to a posterior temperature of $T = 1$ \citep{jaynes2003probability}, Bayesian neural networks have often been observed to achieve better predictive performance when the posterior is artificially cooled with $T < 1$ \citep{osawa2019practicalbayesian, heek2019bayesian, leimkuhler2019partitioned, wenzel2020good}. This CPE behavior is surprising as it suggests that the mathematically correct Bayesian posterior is not always the best-performing posterior in modern neural networks and raises questions about the fundamental modeling assumptions underlying practical Bayesian deep learning.

One possible contributor to the cold posterior effect is approximate inference. Exact posterior inference is computationally intractable for modern neural networks, thus requiring the use of scalable approximate inference methods. There are a wide variety of methods, including, but not limited to, variational inference \citep{blundell2015weightuncertainty}, post-hoc linear Laplace approximations \citep{laplace2021}, and particle-based methods with repulsion in parameter or function space \citep{liu2019svgd, dangelo2023repulsive}. Although these methods make Bayesian inference computationally feasible, they introduce further modeling assumptions that impact predictive performance and uncertainty estimation. For instance, recently \citet{odgers26a} demonstrated that mean-field variational inference methods can overestimate predictive variance and showed that cold posteriors can compensate for this discrepancy. General applications of various scalable inference methods have also found that changing posterior temperatures can often aid in final predictive performance and uncertainty estimation \citep{kapoor2022uncertainty, pitas2023fineprint, shen2024variational}.

Another proposed explanation is prior misspecification. Bayesian neural networks require specifying a prior distribution over the network parameters before observing any data \citep{gelman1996bayesian}. In modern Bayesian neural network training, the standard choice is an isotropic Gaussian prior over the parameters \citep{neal1992bayesian,fortuin2022priorsrevisited}. While computationally convenient, this default choice is not a neutral modeling assumption since the prior directly influences the posterior \citep{mackay1992practical, gelman2017prior, ober2021global, arbel2026primer}. In particular, standard priors may induce unrealistic distributions over functions potentially contributing to the cold posterior effect \citep{wenzel2020good,adlam2020coldposteriors, fortuin2022priorsrevisited}. In fact, \citet{fortuin2022priorsrevisited} showed that priors better matched to the neural network architecture can improve performance and affect the strength of the cold posterior effect. However, much of this evidence has been obtained using computationally intensive sampling-based methods which are often impractical for routine applications of Bayesian neural network training \citep{salimans2015markov, zhang2020cyclicalsgmcmc, kim2024learning, kawasaki2024datasubsampling}. This thus leaves the question about the relative importance of prior choice when using scalable approximate inference algorithms.

Together, these findings motivate an investigation of how approximate inference methods, prior choice, and posterior temperature interact in practice. In this paper, we compare several widely used approximate inference methods across a few common prior families and a range of posterior temperatures on three standard benchmark image classification datasets. Our results suggest that inference method selection and posterior temperature have a substantially greater influence on predictive performance and uncertainty estimation than the choice among simple prior families.

\noindent\textbf{Contributions.} Our main contributions are as follows:
\begin{enumerate} 
\itemsep0em 
\item We show that the choice among simple prior families has little effect on either general metric performance or the shape of the cold posterior distribution in all settings studied.
\item We demonstrate that the choice of approximate method has a substantially greater impact across evaluation metrics on the strength of the cold posterior effect across datasets.
\item We provide practical recommendations for applying scalable Bayesian inference, suggesting the prioritization of inference method selection and temperature tuning over extensive tuning of simple prior families.
\end{enumerate}

\section{Background}
\label{sec:Background}

\subsection{Cold Posterior Effect}

Given data $\mathcal{D}=\{(x_n,y_n)\}_{n=1}^N$ and neural network parameters $\theta \in \mathbb{R}^d$, the standard Bayesian posterior is
\begin{equation}
p(\theta \mid \mathcal{D}) \propto p(\mathcal{D} \mid \theta)p(\theta),
\end{equation}
where $p(\mathcal{D} \mid \theta)$ is the likelihood and $p(\theta)$ is the prior. The temperature parameter is a $T > 0$ such that 
\begin{equation}
p_T(\theta \mid \mathcal{D}) \propto \left[p(\mathcal{D} \mid \theta)p(\theta)\right]^{1/T}.
\end{equation}
The standard Bayes posterior corresponds to $T=1$, while $T<1$ produces a cold posterior that artificially sharpens the posterior distribution. On the other hand, a warm posterior $T > 1$ produces a more diffuse posterior distribution. There exists another formulation known as posterior tempering which applies the temperature parameter only to the likelihood while leaving the prior unchanged. We do not consider this formulation in our study.

In principle, if the prior and likelihood are well specified and inference is exact, the posterior with $T=1$ should be the appropriate Bayesian update \citep{gelman1995bayesian, jaynes2003probability}. However, Bayesian neural networks often achieve better predictive performance when the posterior is artificially cooled with $T<1$ \citep{wenzel2020good}. This observation is surprising because it suggests that the formally correct Bayes posterior may not be the empirically best posterior for deep neural networks. In particular, this suggests that the Bayesian model itself might be misspecified \citep{wenzel2020good, adlam2020coldposteriors}.

\subsection{Approximate Inference}
\label{sec:Approx_Inference}

Exact Bayesian inference in neural networks is generally intractable because the posterior is extremely high-dimensional and often multimodal \citep{mackay1992practical,neal2012bayesian}. As a result, practical Bayesian neural networks rely on approximate inference methods that seek to find a balance between exactness and scalability. In this paper, we compare several such methods, including variational, linear Laplace, and particle-based approaches.

\paragraph{Mean-Field Variational Inference (MFVI).} In general, variational inference methods approximate the exact posterior $p(\theta\mid\mathcal{D})$ by posing a family of distributions and picking from this family the member $q_\phi(\theta)$ that minimizes a divergence with the exact posterior. This is typically done by optimizing the Evidence Lower Bound Objective (ELBO), \begin{equation}
\mathcal{L}_{\mathrm{ELBO}}(\phi) = \mathbb{E}_{q_\phi(\theta)} \left[ \log p(\mathcal{D}\mid \theta) \right]-\mathrm{KL}\!\left(q_\phi(\theta)\,\|\,p(\theta)\right).
\end{equation}
Maximizing this objective is equivalent to minimizing the KL divergence between the variational approximation and the true posterior up to an additive constant \citep{blei2017variational}. 

In mean-field variational inference, the approximate posterior is further assumed to factorize over parameters, typically as a product of independent Gaussian distributions \citep{blei2017variational}, such that 
\begin{equation}
q_\phi(\theta)=\prod_{j=1}^d\mathcal{N}(\theta_j;\mu_j,\sigma_j^2),\qquad\phi = \{\mu,\sigma\}.
\end{equation}
This makes the approximation computationally convenient but it also restricts the posterior family and prevents the approximation from capturing posterior correlations between parameters or non-Gaussian features.

Bayes by Backprop (BBB) \citep{blundell2015weightuncertainty} is one of the standard MFVI methods we consider in this work. BBB learns the parameters of the variational distribution over the weights using stochastic gradient optimization by taking advantage of the reparameterization trick. This makes the algorithm simple and scalable but the mean-field assumptions limit the expressiveness of the posterior approximation.

We also consider Improved Variational Online Newton (IVON), a more recent MFVI method which differs slightly from Bayes by Backprop in how it optimizes the variational distribution \citep{shen2024variational}. IVON is a natural gradient based optimization method that respects the positive definiteness constraints of the covariance matrices in the variational distribution \citep{lin2020handling}. In practice, this gives IVON an optimizer form similar to Adam while still maintaining a Gaussian approximate posterior over weights.

\paragraph{Linear Laplace Approximation.} The linear Laplace Approximation (LLA) is another Gaussian approximation to the posterior. However, unlike the variational methods above, LLA instead builds a local Gaussian approximation around a maximum a posteriori (MAP) estimate \citep{ritter2018scalable, laplace2021}. The MAP estimate $\theta_{\mathrm{MAP}} =\arg\max_\theta\left[\log p(\mathcal{D}\mid \theta)+\log p(\theta)\right]$ is typically found through standard non-Bayesian training methods. The posterior is then approximated with a Gaussian whose covariance is determined by an estimate of the local curvature \citep{tierney1986accurate, laplace2021}
\begin{equation}
q_{\mathrm{LLA}}(\theta)=\mathcal{N}\left(\theta_{\text{MAP}},H_{\text{GGN}}^{-1}\right),
\end{equation}
where $H_{\text{GGN}}$ denotes the diagonal generalized Gauss-Newton (GGN) approximation to the Hessian of the negative log posterior evaluated at the MAP estimate $\left.-\nabla_\theta^2\log p(\theta\mid \mathcal{D})\right|_{\theta=\theta_{\text{MAP}}}$.

In LLA, the model's predictions are additionally linearized with respect to the parameters around the MAP estimate. Specifically, for a neural network $f_\theta$, the prediction is approximated by its first-order Taylor expansion
\begin{equation}
    f_\theta(x) \approx f_{\theta_{\text{MAP}}}(x)+J_{\theta_{\text{MAP}}}(x)(\theta-\theta_{\text{MAP}}),
\end{equation}
where $J_{\theta_{\mathrm{MAP}}}(x)=\nabla_\theta f_\theta(x)\vert_{\theta=\theta_{\text{MAP}}}$ is the Jacobian of the network output with respect to the parameters \citep{immer2021improving}. The predictive distribution over network outputs is also Gaussian with 
\begin{equation}
    q_{\text{LLA}}(f(x) \mid \mathcal{D}) = \mathcal{N}\left(f_{\theta_{\text{MAP}}}(x), J_{\theta_{\text{MAP}}}(x) H_{\text{GGN}}^{-1} J_{\theta_{\text{MAP}}}(x)^\top \right).
\end{equation}
An important note is that in our experiments, we restrict the Laplace approximation to priors for which the implementation supports the required curvature calculations.

\paragraph{Particle-Based Methods.} Particle-based approximate inference methods represent the posterior using a finite set of particles $\{\theta_i\}_{i=1}^M$, where each particle corresponds to a neural network. A simple example is the deep ensemble, which trains multiple networks independently and averages their predictions, often producing strong uncertainty estimates in practice \citep{lakshminarayanan2017simplescalable}. In this paper, we consider two particle-based methods: repulsive ensembles implemented as Wasserstein Gradient Descent (WGD) \citep{dangelo2023repulsive} and Stein Variational Gradient Descent (SVGD) \citep{liu2019svgd}.

SVGD derives its update rule from Stein variational inference by evolving a set of particles toward the target posterior using an update that combines an attractive term toward high posterior density regions with a kernel-based repulsive term that encourages particle diversity \citep{liu2019svgd}. Explicitly, we have
\begin{equation}\theta_i^{t+1}=\theta_i^t+\eta_t\phi_{\mathrm{SVGD}}(\theta_i^t), \text{ where }\phi_{\mathrm{SVGD}}(\theta_i)=\frac{1}{M}\sum_{j=1}^M\left[k(\theta_j,\theta_i) \nabla_{\theta_j} \log p(\theta_j\mid \mathcal{D})+\nabla_{\theta_j}k(\theta_j,\theta_i)\right].
\end{equation}
The first term, $k(\theta_j,\theta_i)\nabla_{\theta_j}\log p(\theta_j \mid \mathcal{D})$, moves particles toward regions of high posterior density, while the second term, $\nabla_{\theta_j}k(\theta_j,\theta_i)$, provides a kernel-based repulsive force that discourages particle collapse \citep{liu2019svgd}.

WGD, on the other hand, can be written as a particle approximation to the Wasserstein gradient flow of the KL divergence \citep{dangelo2023repulsive}. If $\rho_t$ denotes the current particle distribution, the idealized update is
\begin{equation}\theta_i^{t+1}=\theta_i^t+\eta_t\left[\nabla_{\theta_i^t}\log p (\theta_i^t \mid \mathcal{D})-\nabla_{\theta_i^t} \log \rho_t(\theta_i^t)\right].
\end{equation}
The first term again moves particles toward high posterior density, while the second term moves particles away from regions where the particle density is already high. Since $\rho_t$ is unknown, it is commonly estimated with a kernel density estimator, giving
\begin{equation}\nabla_{\theta_i^t} \log \rho_t(\theta_i^t)\approx\frac{\sum_{j=1}^M\nabla_{\theta_i^t} k(\theta_i^t,\theta_j^t)}{\sum_{j=1}^Mk(\theta_i^t,\theta_j^t)}.
\end{equation}
In our experiments, we use the radial basis function (RBF) kernel for both SVGD and WGD, defined by 
$k(\theta_i,\theta_j) =\exp\left(-\frac{1}{h}\|\theta_i-\theta_j\|^2\right)$
where $h$ denotes the lengthscale.

Intuitively, although both methods combine attraction toward high posterior density regions with repulsion between particles, they differ in how this particle motion is defined. SVGD uses a kernelized Stein gradient, while WGD follows the Wasserstein gradient flow of the KL divergence \citep{liu2019svgd, dangelo2023repulsive}. In practice, SVGD averaging posterior gradients across particles can become problematic in high-dimensional parameter spaces, while WGD avoids this averaging and may be preferable for high-dimensional neural network ensembles \citep{dangelo2023repulsive}.

\paragraph{Function-Space Particle Methods.} The particle-based methods above can also be formulated in function space, where diversity is encouraged between the functions represented by the particles rather than directly between their parameters \citep{wang2019functionspaceparticleoptimization}. Neural networks are highly overparameterized and very different parameters can potentially represent similar predictive functions. As a result, diversity in weight space does not necessarily imply diversity in the induced predictions. Function-space particle methods thus define repulsion over network outputs rather than over weights, so the particles are pushed apart according to their predictive behavior instead of their location in parameter space \citep{dangelo2023repulsive}. 

\section{Related Works}

Several explanations for the cold posterior effect have been proposed. One prominent line of work attributes the effect to likelihood misspecification \citep{zeno2020cold, aitchison2021statistical, noci2021disentangling, kapoor2022uncertainty}. Standard Bayesian classifiers often use a categorical likelihood, but image classification datasets are typically shaped by collection, filtering, labeling, cleaning, and data augmentation. In particular, \citet{izmailov2021bayesian} found that the cold posterior effect in their experiments was largely a result of standard data augmentation practices. However,  \citet{nabarro2021dataaugmentation} show that these effects can persist even under principled Bayesian treatments of augmentation, suggesting that data augmentation alone does not explain the phenomenon. We do not explore likelihood misspecification or data augmentation as explanations for the cold posterior effect in this paper. In particular, we omit data augmentation from our experiments to avoid this potential confounding factor when studying prior choice and approximate inference.
 
Another explanation is prior misspecification \citep{wenzel2020good, fortuin2022priorsrevisited}. Most Bayesian neural networks use simple isotropic Gaussian priors over parameters even though these priors may not match the structure of trained neural networks or their induced functions \citep{neal1992bayesian, fortuin2022priorsrevisited}. Because neural networks are highly nonlinear, even simple weight-space priors can accidentally induce complex and unintuitive distributions over the generated functions \citep{tran2022functionalpriors,fortuin2022priors}. In particular, \citet{fortuin2022priorsrevisited} find that unfit priors which place too much mass on implausible parameters for some particular neural network architecture can affect both predictive performance and the strength of the cold posterior effect.

Approximate inference has also been proposed as a possible contributor to the cold posterior effect \citep{wenzel2020good}. Bayesian neural network posteriors are high-dimensional, complex, and difficult to approximate, so approximate inference methods may introduce method-specific biases \citep{blundell2015weightuncertainty, izmailov2021bayesian}. In particular, \citet{izmailov2021bayesian} show that conclusions about cold posteriors can change under more accurate full-batch Hamiltonian Monte Carlo inference. Thus, the cold posterior effect may depend not only on the specified Bayesian model, but also on how the posterior is approximated in practice \citep{odgers26a}.

Motivated by these explanations, this paper studies both prior choice and approximate inference. If prior misspecification is a primary driver of the cold posterior effect, then replacing the default Gaussian prior with alternative priors should change the observed cold posterior effect across different approximate inference methods. If, instead, cold posterior behavior varies more strongly across inference algorithms than across priors, then practical inference might play a larger role than the prior in determining how and when the cold posterior effect emerges.

\section{Experimental Setup}

Building on the experimental framework of \citet{fortuin2022priorsrevisited}, we performed experiments on three standard image classification benchmarks: MNIST, FashionMNIST, and CIFAR-10. We use CNNs for MNIST and FashionMNIST and ResNet18 \citep{he2015deepresiduallearningimage} for CIFAR-10. For each approximate inference method described in Section \ref{sec:Approx_Inference}, we repeat the experiments under four prior distributions: Gaussian, Laplace, Cauchy, and Student-t. The prior hyperparameters are fixed across experiments so that differences in performance can be attributed primarily to the interactions between inference method, prior family, and posterior temperature rather than to method-specific prior tuning.

To study the cold posterior effect, we logarithmically vary the posterior temperature over the range $T = 1$ to $T=10^{-7}$, where $T=1$ corresponds to the standard Bayesian posterior. This temperature sweep is applied separately for every combination of dataset, approximate inference method, and prior distribution.

We note that while data augmentation is standard when training ResNets on CIFAR-10 \citep{he2015deepresiduallearningimage}, we do not use data augmentation in our experiments. This is because prior work has shown that data augmentation can interact nontrivially with Bayesian inference and posterior tempering \citep{izmailov2021bayesian, nabarro2021dataaugmentation}. Since our goal is to isolate how approximate inference method and prior choice affect the cold posterior effect, we omit data augmentation in order to avoid introducing augmentation as an additional factor.

For each experimental configuration, we perform a small hyperparameter grid search over learning rate, learning rate halving, and batch size centered around the settings recommended in the respective papers introducing each inference method, the specific details of which are listed in Appendix \ref{sec:hyperparameters}. Hyperparameters are selected by choosing the configuration that achieves the highest validation accuracy. For each dataset, inference method, prior, and temperature, we run three independent replicates with different random seeds and report the mean performance with standard deviations across seeds.

We compare all Bayesian methods against the standard SGD baseline with weight decay, which acts as a MAP approximation with a Gaussian prior \citep{bishop2006pattern}. The SGD baseline is included as a reference point to assess how competitive the Bayesian methods are relative to standard deterministic training. This comparison is particularly important in the cold posterior setting, where previous work has observed that the standard Bayesian posteriors can underperform these deterministic baselines unless the posterior is artificially cooled \citep{wenzel2020good}. We evaluate each method using four metrics: accuracy, negative log-likelihood (NLL), expected calibration error (ECE), and root Brier score (RBS). The code for the experiments is available at \url{https://github.com/duckpy/practical-bnn-priors}.

\section{Experimental Results}

In this section, we highlight the two primary findings from the experiments. First, we find that when using approximate inference, the choice amongst simple prior families has little to no effect. Second, we find that the strength of the observed CPE is strongly dependent on the inference method. More generally, performance across metrics varies mostly as a function of the choice of inference method.

\subsection{Weak Prior Dependence in Approximate Inference}
\label{sec:CPE_Prior}

\begin{figure}[ht]
\centering
\includegraphics[width=0.87\linewidth]{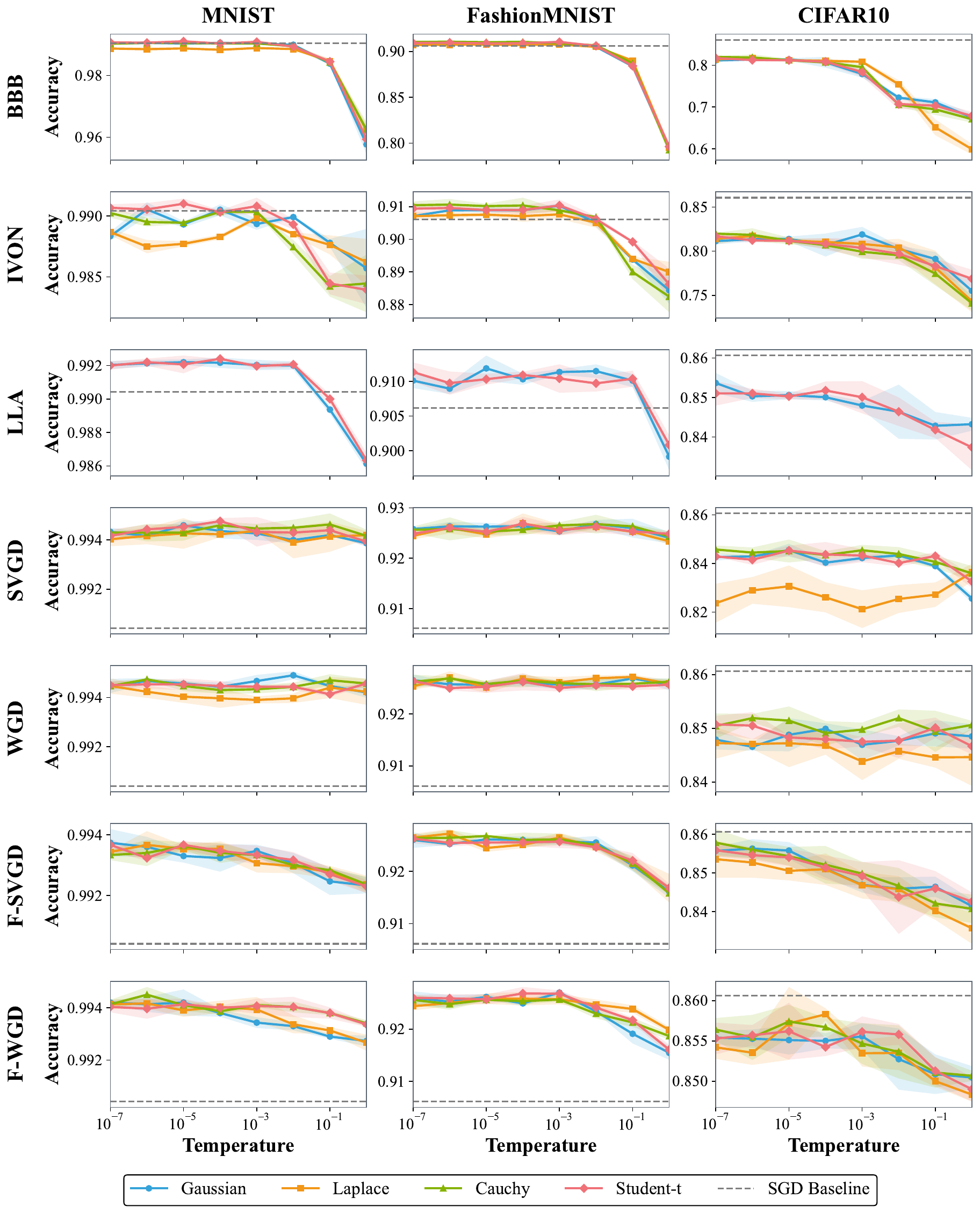}
\caption{Accuracy as a function of temperature for MNIST, FashionMNIST, and CIFAR10. Rows correspond to inference methods and columns to datasets. Within each panel, we compare Gaussian, Laplace, Cauchy, and Student-t priors with the dashed line indicating the SGD baseline. In general, prior choice has little to no effect on accuracy while CPE behavior and accuracy performance vary across inference methods. Higher is better.}
\label{fig:acc}
\end{figure}

Across datasets and evaluation metrics, we find that the choice among the simple prior families considered has little effect on either predictive performance or temperature-dependent behavior. This is illustrated clearly by the accuracy results in Figure \ref{fig:acc}. Within a fixed inference method and dataset, the curves associated with different prior families have significant overlap and follow closely aligned trajectories across temperature. Even when considering distributions with substantially different tail behavior, we observe no consistent advantage from substituting in one prior family for another.

These findings provide an interesting contrast to prior work investigating the role of prior specification in Bayesian neural networks. In particular, \citet{fortuin2022priorsrevisited} demonstrated that heavy-tailed priors can improve predictive performance and mitigate the cold posterior effect using stochastic-gradient MCMC. However, their supplementary experiments with MFVI found little sensitivity to prior choice. Our results thus extend this observation across a broader range of practical inference methods, suggesting that the influence of prior specification may depend substantially on the inference procedure used to approximate the posterior.

It is important to note that these results do not imply that prior specification is unimportant in Bayesian neural networks. Rather, we find that when using the approximate inference methods considered here, differences among simple prior families have a limited influence on predictive behavior. The corresponding NLL, ECE, and RBS results are provided in Appendix \ref{sec:unc_metrics}. We omit these figures from the main text because they do not reveal observations beyond those evident in the accuracy results.

\subsection{Overall Performance Across Different Inference Methods}
\label{sec:Overall_Performance}

Having established that simple prior family choice produces comparatively small differences, we next examine the effect of the approximate inference method itself. We focus on the Gaussian prior in the main text and report the corresponding results for the remaining priors in Appendix \ref{sec:prior}. As discussed previously, the conclusions are largely unchanged across prior families.

\begin{figure}[H]
\centering
\includegraphics[width=0.95\linewidth]{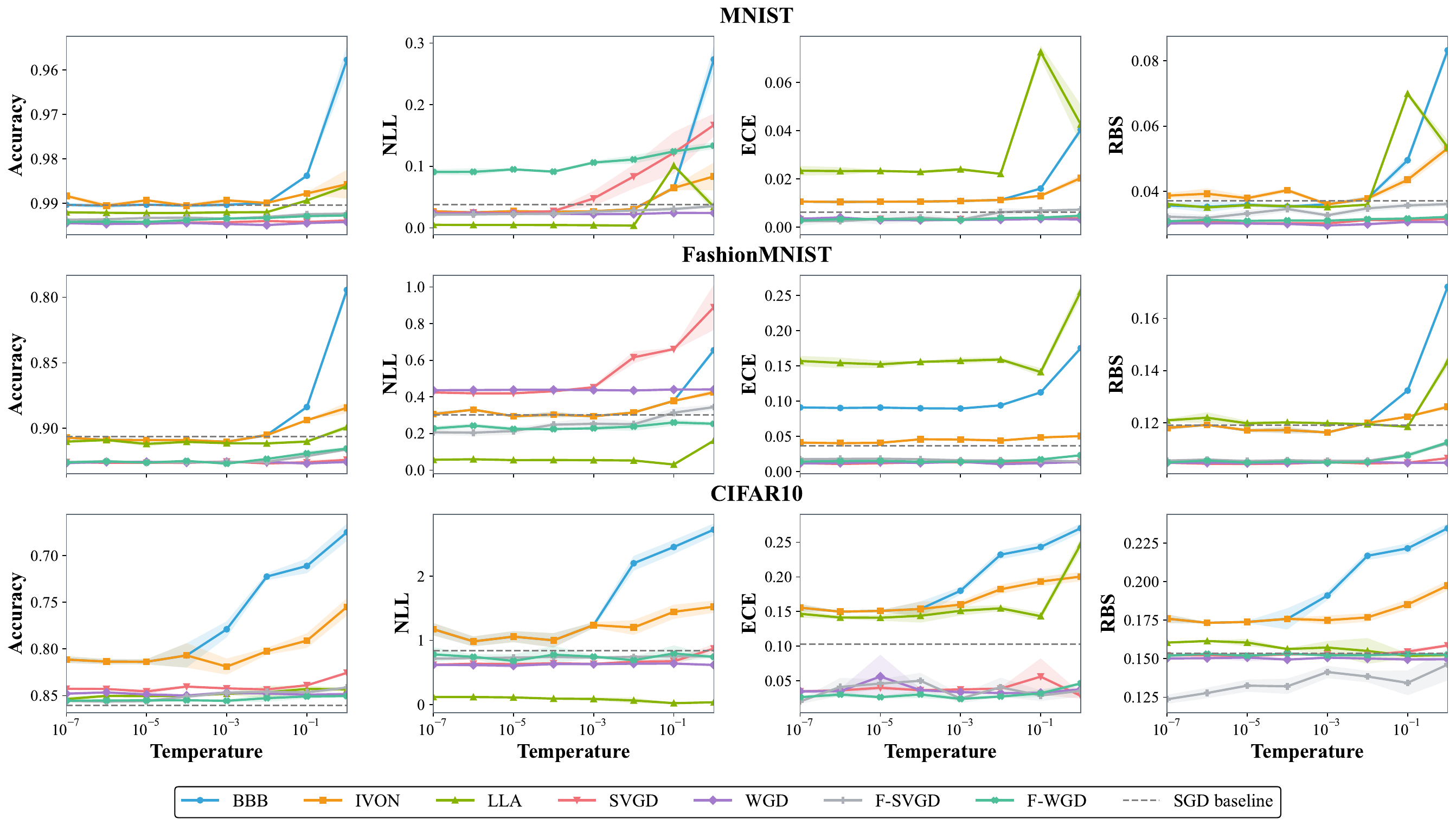}
\caption{Performance of approximate Bayesian inference methods under a fixed Gaussian prior across posterior temperatures. All panels are oriented so that lower is better. Across datasets, cold temperatures generally improve or preserve performance relative to warmer temperatures, while ensemble and particle-based methods, especially WGD and the function-space variants, provide the most stable uncertainty-aware predictions.}
\label{fig:gaussian}
\end{figure}

Figure \ref{fig:gaussian} reveals two main conclusions as we vary the approximate inference method used. First, the strength of the cold posterior effect varies substantially across inference methods. These method-specific differences persist across the datasets considered, although the magnitudes of the effect do vary by dataset. 

Specifically, the Gaussian approximation methods such as BBB, IVON, and the LLA display a pronounced CPE in both predictive and uncertainty metrics. One possible explanation comes from \citet{odgers26a}, who demonstrate that posterior cooling can partially correct the overestimation of predictive variance and the underestimation of parameter-space posterior variance in MFVI. Although their analysis focuses on MFVI, it perhaps suggests that the discrepancies caused by Gaussian approximation methods may be a reason for the observed temperature sensitivity.

In contrast, we find that the particle-based methods exhibit weaker temperature dependence overall, with function-space particle methods displaying a slight temperature dependence while parameter-space particle methods remain relatively stable across temperatures. We believe this slight difference between parameter-space and function-space methods is due to the fact that under parameter-space methods might experience functional distribution collapse. That is, if two or more particles end up in different high posterior density regions, function-space repulsion will help particles spread out across these regions to better approximate the local shape of the posterior predictive distribution around the maxima. As a result, posterior temperature can have a larger effect as sharpening the posterior changes how strongly particles concentrate around the maxima.

Second, absolute performance across metrics also varies considerably with the inference method. Across the three datasets, the MFVI methods BBB and IVON consistently have the weakest performance even in their best temperature ranges. For accuracy and RBS, these MFVI methods are followed closely by the LLA. However, the LLA exhibits substantially greater metric dependence than the other approximate inference methods with strong performance in NLL while performing considerably worse in ECE across datasets. In contrast, the particle-based methods consistently exhibit strong performance across datasets and metrics, often matching or exceeding the SGD baseline.

For these strongest performing particle methods, we observe that lower temperatures typically perform comparably to, and in several cases slightly better than, the standard $T = 1$. Overall, the results suggest that those seeking a scalable Bayesian inference method should first consider ensemble-based approximations, especially WGD or its function-space variant with a low posterior temperature.

\section{Discussion \& Conclusion}

This paper studied how prior choice, posterior temperature, and approximate inference method interact in practical Bayesian neural networks. Across several datasets, metrics, and inference algorithms, we found that inference method and posterior temperature usually have a larger and more consistent effect on performance than the choice among simple weight-space prior families. The cold posterior effect is also strongly method-dependent as it is pronounced for mean-field variational methods, mixed for Laplace approximation, weaker for parameter-space particle methods, and more visible again for function-space particle methods. These results suggest that the type of approximate inference plays a central role in determining when and how posterior tempering improves performance.

These findings should not be interpreted as evidence that priors are unimportant in Bayesian neural networks. Our experiments test a limited set of simple weight-space prior families with fixed hyperparameters. Looking deeper into structured priors \citep{karaletsos2020hierarchical,pearce2020structured}, architecture-aware priors \citep{fortuin2022priorsrevisited}, functional priors \citep{tran2022functionalpriors}, or learned priors \citep{sam2024bayesian, rochussen2026amortisinginferencemetalearningpriors}, one could find different effects. However, for the purpose of practical scalable approximate inference, our results suggest that simple prior-family choice is typically less important than the choice of inference method and temperature.

It is also important to consider the computational constraints of our experimental setup. The hyperparameter tuning budget was limited, and thus the reported results may not represent the best possible performance of each method. At the same time, however, our tuning procedure was intended to reflect practical settings in which exhaustive hyperparameter optimization might be impractical. Furthermore, in our experiments, hyperparameters were selected based on validation accuracy, which may favor settings that improve classification performance rather than uncertainty-sensitive metrics.

For practitioners seeking to use practical BNNs, the main recommendation suggested by our results is to prioritize approximate inference method selection and to allocate subsequent tuning effort to posterior temperature before investing resources in the choice between simple prior families. More broadly, our findings indicate that conclusions about the benefits of posterior tempering cannot be separated from the approximate inference method used to obtain the posterior approximation.

\newpage

\subsubsection*{Acknowledgments}
VF was supported by the Branco Weiss
Fellowship.

\bibliography{main}
\bibliographystyle{tmlr}

\newpage
\appendix
\section{Training Details and Hyperparameters}
\label{sec:hyperparameters}

Here, we detail the exact hyperparameter grids used for each inference method. The grid search was independently executed for each combination of inference method, dataset, prior, and temperature. 

The search ranges for each hyperparameter were informed by values reported in previous applications of the respective methods. Specifically, the hyperparameter grids for BBB, IVON, LLA, SVGD, WGD, F-SVGD, and F-WGD were informed by \citet{blundell2015weightuncertainty}, \citet{shen2024variational}, \citet{laplace2021} and \citet{immer2021improving}, \citet{liu2019svgd}, \citet{dangelo2023repulsive}, \citet{wang2019functionspaceparticleoptimization}, and \citet{dangelo2023repulsive} respectively. While learning rate halving was used in some of these previous applications, it was not universally applied. In these cases, the learning rate schedules were hand selected by us.

\begin{figure}[h]
\begin{center}
\begin{tabular}{ | >{\centering\arraybackslash}m{8em} | >{\centering\arraybackslash}m{9em}| >{\centering\arraybackslash}m{8em} | >{\centering\arraybackslash}m{8em} | } 
  \hline
 \textbf{Method} & \textbf{LR} & \textbf{LR Halving}& \textbf{Batch Size}  \\ 
  \hline
  BBB & $10^{-2},10^{-3},10^{-4}$& None, 50, 100 & 32, 128 \\ 
  \hline
   IVON& $10^{-1},10^{-2}$ & None, 50, 100 & 32, 128 \\ 
  \hline
  LLA & $10^{-1},10^{-2}$ & None, 25 & 32, 128 \\ 
  \hline
  SVGD& $10^{-2},10^{-3},10^{-4}$ & None, 50 & 32, 128 \\ 
  \hline
  WGD& $10^{-2},10^{-3}$ & None, 25, 50 &  32, 128, 256\\ 
  \hline
  F-SVGD& $10^{-3},10^{-4}$ & None, 25, 50 &  32, 128\\ 
  \hline
  F-WGD & $10^{-2},10^{-3}$ & None, 25, 50 & 32, 128, 256\\ 
  \hline
\end{tabular}
\end{center}
\caption{Hyperparameter search spaces for each inference method. LR halving denotes the interval, in epochs, at which the learning rate is multiplied by $1/2$, with ``None'' indicating a constant learning rate.}
\end{figure}

\section{Additional Uncertainty Metrics}
\label{sec:unc_metrics}

\begin{figure}[H]
\centering
\textbf{NLL Across Experiments}\par\medskip\hspace*{-0.65cm}
\includegraphics[width=0.85\linewidth]{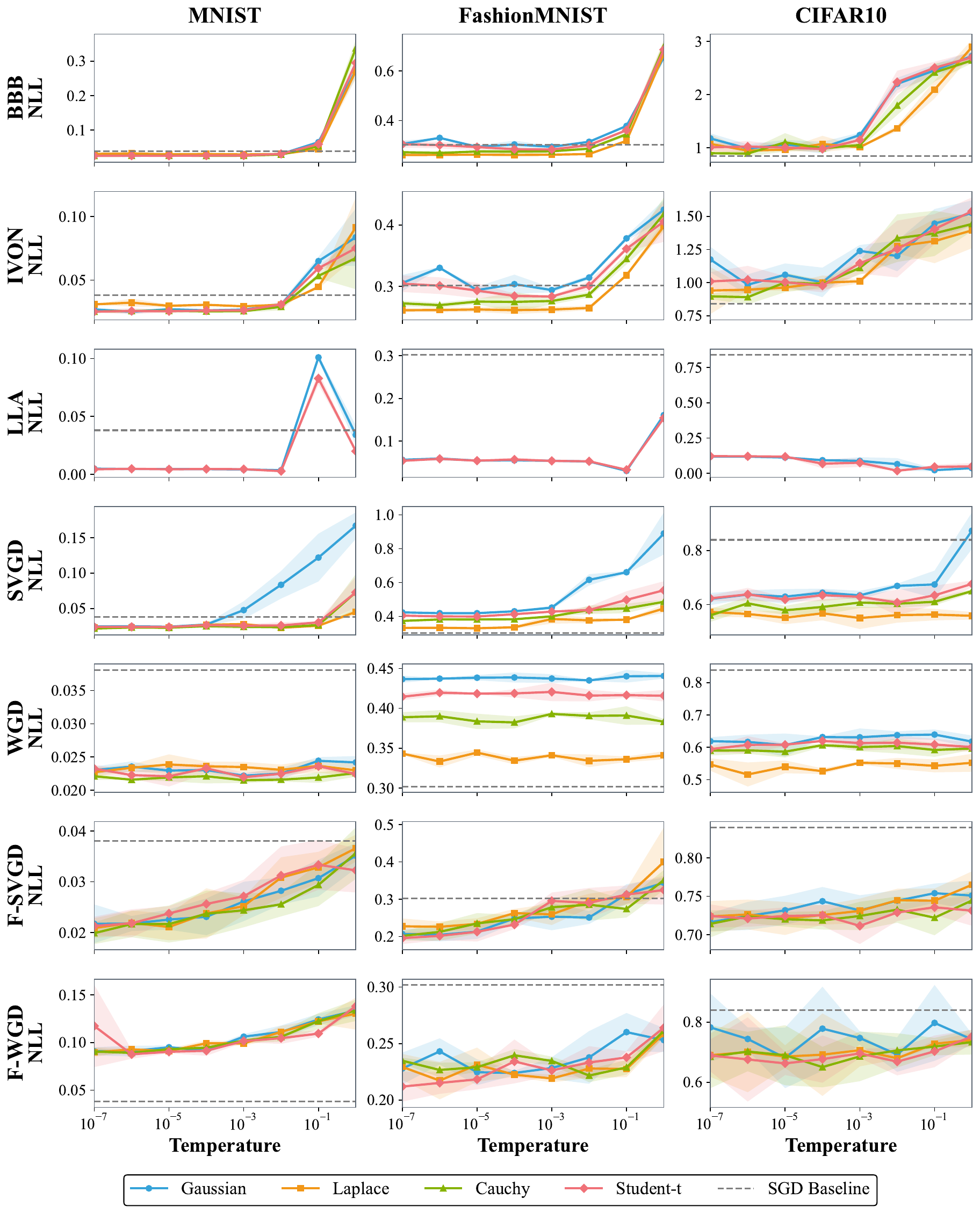}
\caption{NLL as a function of temperature for MNIST, FashionMNIST, and CIFAR10. Rows correspond to inference methods and columns to datasets. Within each panel, we compare Gaussian, Laplace, Cauchy, and Student-t priors with the dashed line indicating the SGD baseline. We attribute the artifact for the spike in the LLA panels to the hyperparameter tuning being done based on accuracy. Furthermore, we observe that for some inference method and dataset combinations, there is some separation between the prior families. However, the scale of this separation is still much less than the separation across inference methods.}
\label{fig:nll}
\end{figure}

\begin{figure}[H]
\centering
\textbf{ECE Across Experiments}\par\medskip\hspace*{-0.65cm}
\includegraphics[width=0.85\linewidth]{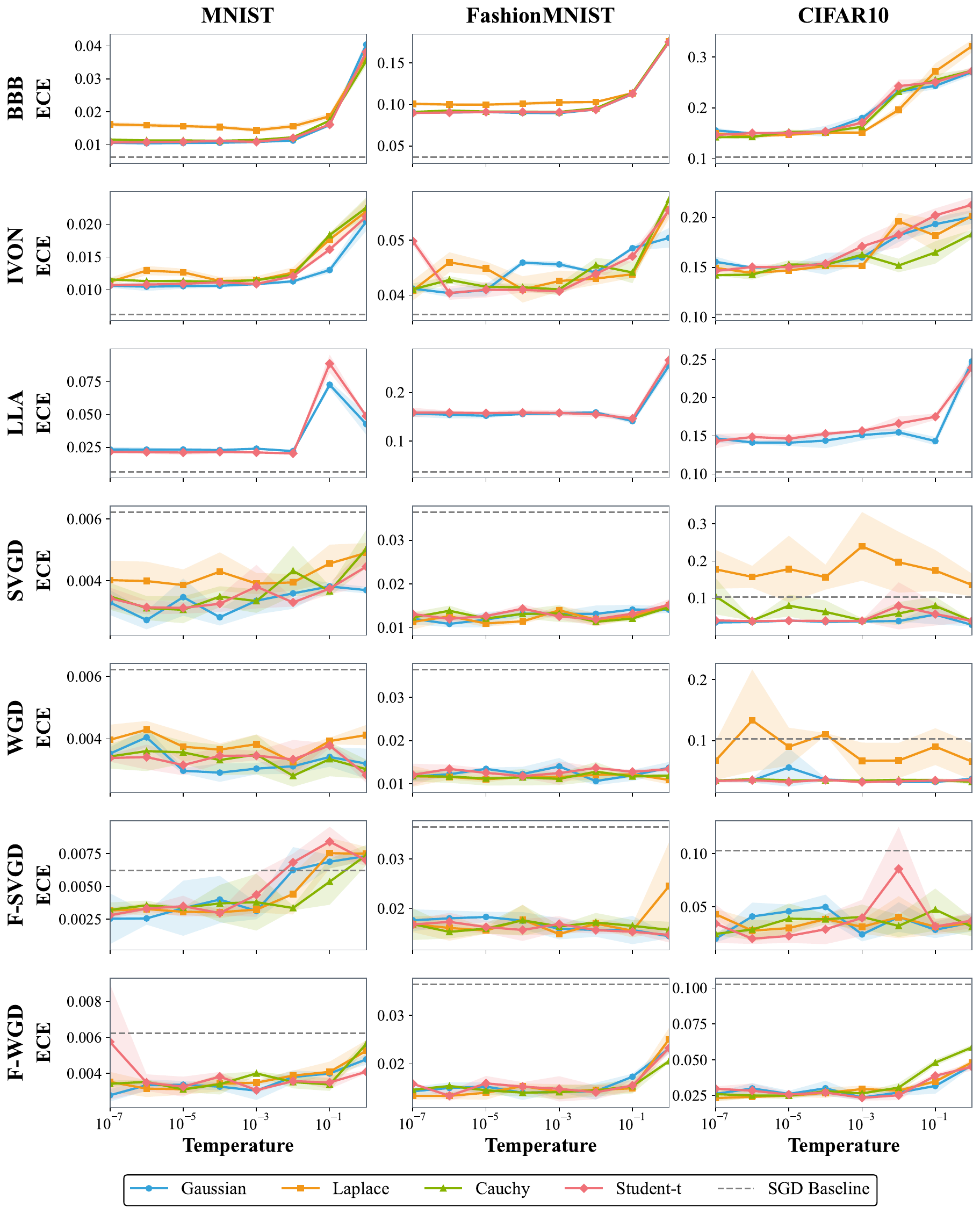}
\caption{ECE as a function of temperature for MNIST, FashionMNIST, and CIFAR10. Rows correspond to inference methods and columns to datasets. Within each panel, we compare Gaussian, Laplace, Cauchy, and Student-t priors with the dashed line indicating the SGD baseline. We again attribute the artifact for the spike in the LLA panels to the hyperparameter tuning being done based on accuracy.}
\label{fig:ece}
\end{figure}

\begin{figure}[H]
\centering
\textbf{RBS Across Experiments}\par\medskip\hspace*{-0.65cm}
\includegraphics[width=0.85\linewidth]{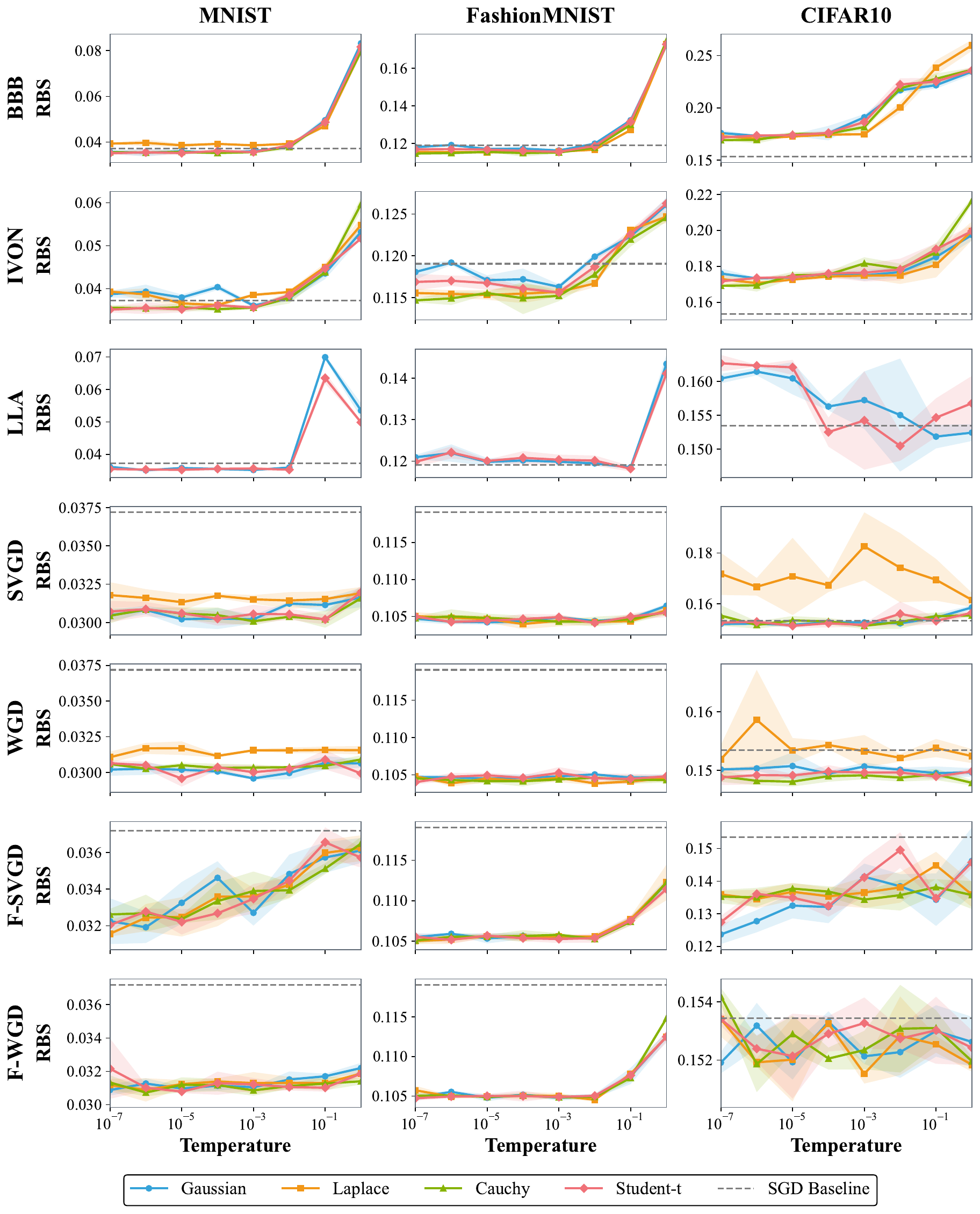}
\caption{RBS as a function of temperature for MNIST, FashionMNIST, and CIFAR10. Rows correspond to inference methods and columns to datasets. Within each panel, we compare Gaussian, Laplace, Cauchy, and Student-t priors with the dashed line indicating the SGD baseline. We again attribute the artifact for the spike in the LLA panels to the hyperparameter tuning being done based on accuracy.}
\label{fig:rbs}
\end{figure}

We present the figures for the uncertainty metrics NLL, ECE, and RBS that correspond with the accuracy plots in Figure \ref{fig:acc} of Section \ref{sec:CPE_Prior}. Similarly to what was observed for the predictive performance in Section \ref{sec:CPE_Prior} we find little difference in metric performance when varying between the tested prior families in Figures \ref{fig:nll}, \ref{fig:ece}, and \ref{fig:rbs}. We do note, however, that for some combinations of dataset and inference method, there is some separation between the prior curves. We observe this in particular for WGD and FashionMNIST. However, the difference between these prior curves is much less significant compared to the differences between approximate inference methods.

\newpage

\newpage

\section{Additional Fixed Prior Figures}
\label{sec:prior}

\begin{figure}[H]
\centering
\textbf{Laplace Prior}\par\medskip\hspace*{-0.65cm}
\includegraphics[width=0.83\linewidth]{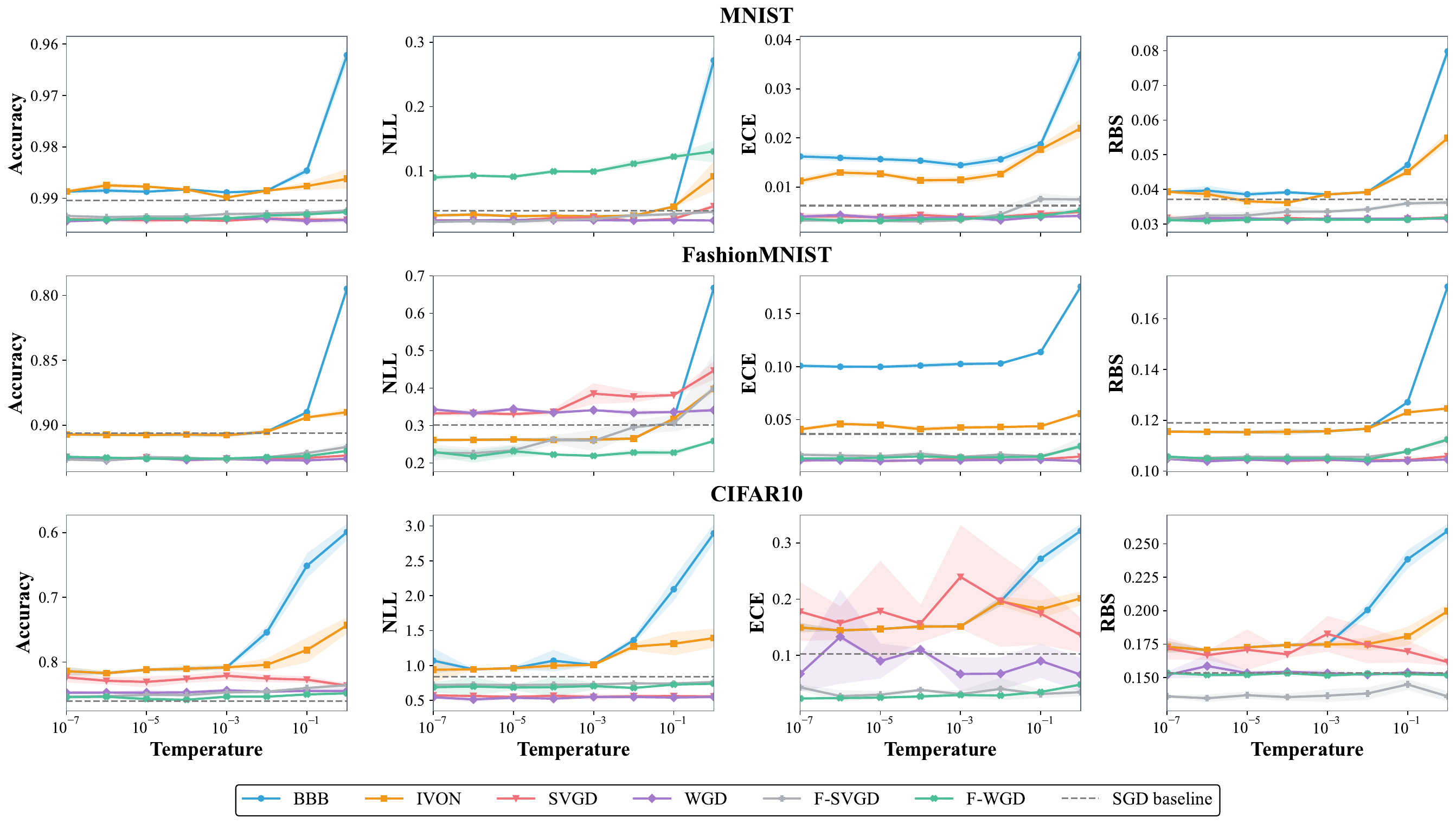}
\caption{Performance of approximate Bayesian inference methods under a fixed Laplace prior across posterior temperatures. All panels are oriented so that lower is better.}
\label{fig:laplace}
\end{figure}

\begin{figure}[H]
\centering
\textbf{Cauchy Prior}\par\medskip\hspace*{-0.65cm}
\includegraphics[width=0.83\linewidth]{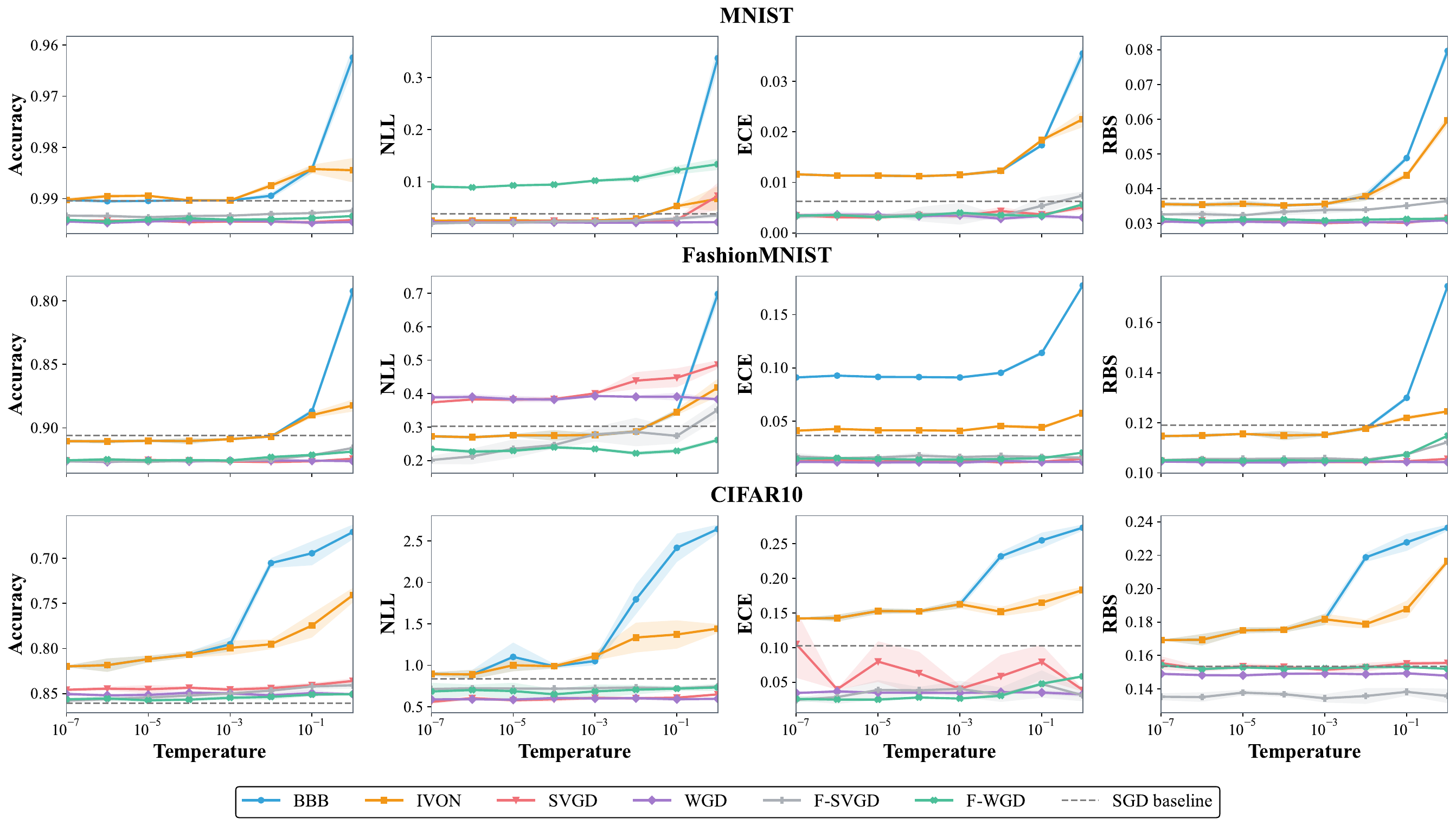}
\caption{Performance of approximate Bayesian inference methods under a fixed Cauchy prior across posterior temperatures. All panels are oriented so that lower is better.}
\label{fig:cauchy}
\end{figure}

\begin{figure}[H]
\centering
\textbf{Student-t Prior}\par\medskip\hspace*{-0.65cm}
\includegraphics[width=0.83\linewidth]{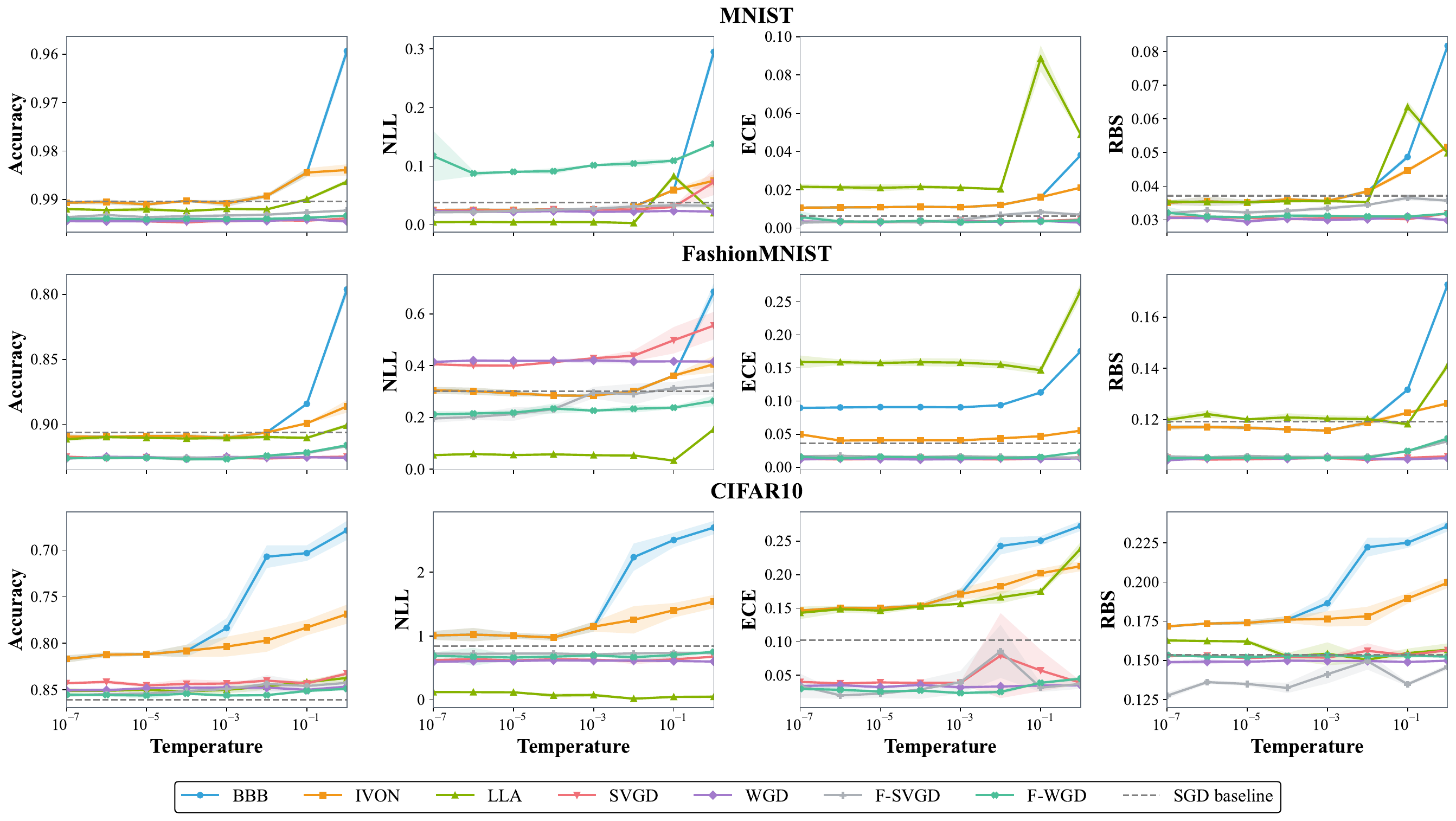}
\caption{Performance of approximate Bayesian inference methods under a fixed Student-t prior across posterior temperatures. All panels are oriented so that lower is better.}
\label{fig:studentt}
\end{figure}

We present the fixed prior figures for the Laplace, Cauchy, and Student-t priors that correspond to the Gaussian prior plots in Figure \ref{fig:gaussian} of Section \ref{sec:Overall_Performance}. We observe the same pattern with the Gaussian prior across Figures \ref{fig:laplace}, \ref{fig:cauchy}, and \ref{fig:studentt}. Specifically, the strength of the cold posterior effect varies substantially across inference methods with absolute performance varying considerably with inference method choice.

\end{document}